\documentclass[preprint,12pt]{elsarticle}

\usepackage{amssymb}
\usepackage{amsmath}
\usepackage{url}
\usepackage{svg}
\usepackage{array}
\usepackage{booktabs}
\usepackage{multirow}
\usepackage[table]{xcolor}
\usepackage[linesnumbered,ruled,vlined]{algorithm2e}
\usepackage{algpseudocode}

\journal{Computer Vision and Image Understanding}

\begin{document}

\makeatletter
\def\ps@pprintTitle{%
  \let\@oddhead\@empty
  \let\@evenhead\@empty
  \let\@oddfoot\@empty
  \let\@evenfoot\@oddfoot}
\makeatother

\begin{frontmatter}



\title{T-SiamTPN: Temporal Siamese Transformer Pyramid Networks for Robust and Efficient UAV Tracking} 

%
\author[1]{Hojat Ardi} 
\author[1]{Amir Jahanshahi\corref{cor1}}
\cortext[cor1]{Corresponding author}
\ead{amir.jahanshahi@aut.ac.ir}
\author[2]{Ali Diba\corref{cor1}}
\ead{adiba@hbku.edu.qa}

\affiliation[1]{
              organization={Department of Electrical Engineering, Amirkabir University of Technology (AUT)}, 
              city={Tehran},
              country={Iran}
}

\affiliation[2]{
              organization={Qatar Computing Research Institute Hamad Bin Khalifa University}, 
              city={Doha},
              country={Qatar}
}




\begin{abstract}
Aerial object tracking remains a challenging task due to scale variations, dynamic backgrounds, clutter, and frequent occlusions. While most existing trackers emphasize spatial cues, they often overlook temporal dependencies, resulting in limited robustness in long-term tracking and under occlusion. Furthermore, correlation-based Siamese trackers are inherently constrained by the linear nature of correlation operations, making them ineffective against complex, non-linear appearance changes.
To address these limitations, we introduce T-SiamTPN, a temporal-aware Siamese tracking framework that extends the SiamTPN architecture with explicit temporal modeling. Our approach incorporates temporal feature fusion and attention-based interactions, strengthening temporal consistency and enabling richer feature representations. These enhancements yield significant improvements over the baseline and achieve performance competitive with state-of-the-art trackers.
Crucially, despite the added temporal modules, T-SiamTPN preserves computational efficiency. Deployed on the resource-constrained Jetson Nano, the tracker runs in real time at 7.1 FPS, demonstrating its suitability for real-world embedded applications without notable runtime overhead.
Experimental results highlight substantial gains: compared to the baseline, T-SiamTPN improves success rate by 13.7\% and precision by 14.7\%. These findings underscore the importance of temporal modeling in Siamese tracking frameworks and establish T-SiamTPN as a strong and efficient solution for aerial object tracking.
Code is available at: 
\url{https://github.com/to/be/released}
\end{abstract}




\end{frontmatter}



\section{Introduction}
\label{sec1:intro}
Visual object tracking is a fundamental and active research area within the field of computer vision, aiming to estimate and continuously update the position of one or more objects of interest as they move through a video sequence. This task involves not only detecting objects in individual frames but also maintaining their identities over time, even in the presence of challenges such as occlusion, scale variation, illumination changes, fast motion, and background clutter.

Due to its broad utility, visual tracking has become a critical component in numerous real-world applications. For instance, in autonomous driving systems, robust tracking is essential for understanding the dynamics of surrounding vehicles and pedestrians. In video surveillance, tracking supports abnormal activity detection and real-time monitoring. Other key application domains include augmented and virtual reality, human-computer interaction, robotics, and military systems, where accurate tracking contributes to situational awareness and decision-making \cite{xing2021visual}.

Tracking algorithms are generally categorized into two principal types: Single-Object Tracking (SOT) and Multiple-Object Tracking (MOT). In the SOT paradigm, the tracker is given an initial bounding box of the target in the first frame and is required to follow that specific target across subsequent frames, without relying on pre-trained object detectors or labeled training data \cite{ma2022unified}. This scenario is often used in real-time applications where immediate tracking is needed after manual initialization.

In contrast, MOT is a more complex task that involves detecting and tracking multiple targets simultaneously throughout a video sequence \cite{meinhardt2022trackformer}. It requires solving both the detection problem and the data association problem — ensuring that each detection is correctly assigned to an existing trajectory or initialized as a new one.

\begin{figure}[t]
    \centering
   \includegraphics[width=1\columnwidth]{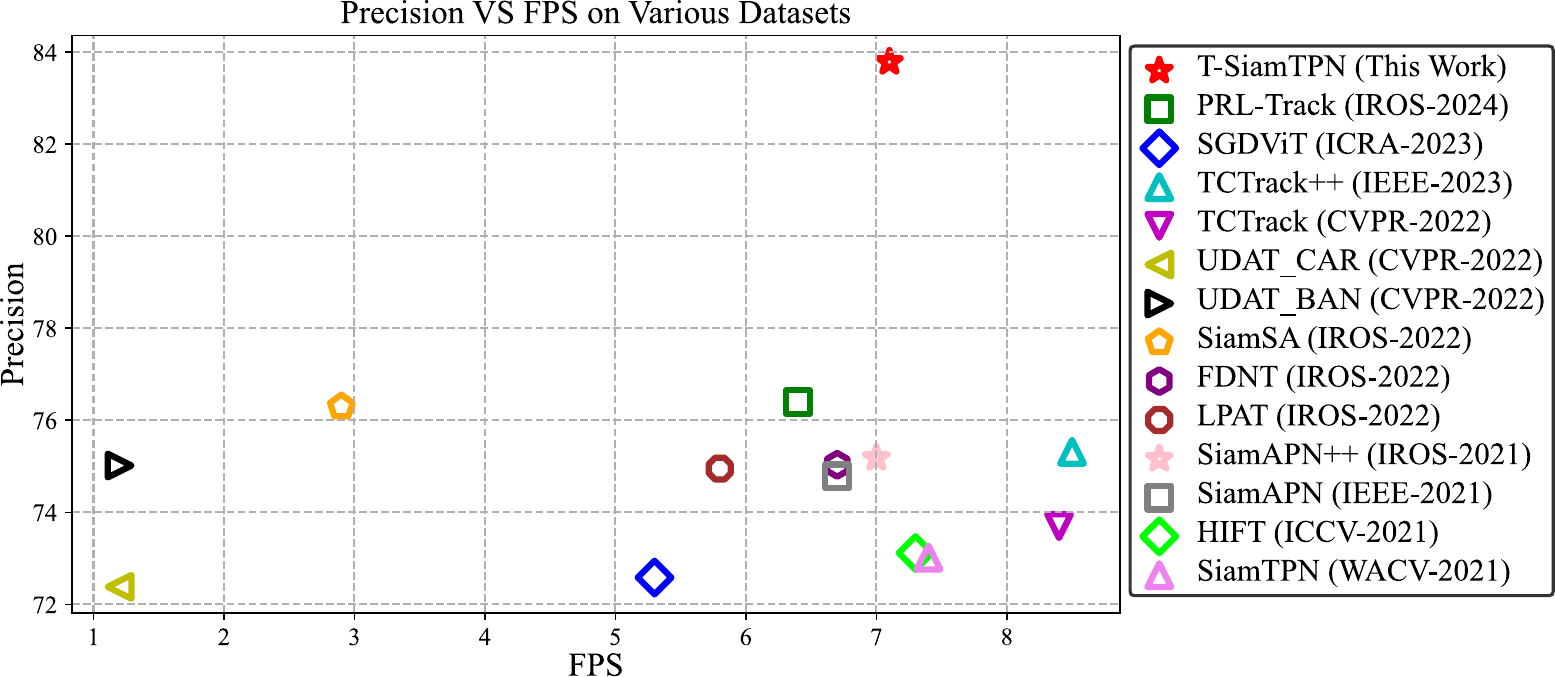} 
    \caption{Comparison of T-SiamTPN with state-of-the-art aerial trackers in terms of precision and speed. The proposed model (red star) achieves a well-balanced trade-off between performance and efficiency, attaining a high average precision of 83.8\% and a processing speed of 7.1 FPS.}
    \label{fig:1}
\end{figure}

UAV-based aerial tracking has gained significant attention due to its wide field of view and high flexibility, particularly in applications like surveillance and aerial imaging. However, tracking in aerial videos faces numerous challenges, including small target sizes, fast motions, perspective changes, occlusions, cluttered backgrounds, objects resembling the target, and computational limitations in embedded systems. Since UAVs typically rely on lightweight, low-power hardware, they struggle to execute computationally intensive algorithms. This creates a balancing act between tracking performance, processing speed, and energy efficiency, especially in real-time scenarios. 

To address all these issues, several visual tracking algorithms have been developed, and most are feature-based and deep learning-based \cite{kugarajeevan2023transformers}. Traditional feature-based algorithms  such as \cite{liu2011robust, zhang2013human, bertinetto2016staple, li2013survey, kokul2015online} are preprogrammed-rule-based and fail to perform in cases with occl, usions, background complexity, or motion with a high speed. Deep learning-based algorithms, i.e., convolutional neural networks (CNN) and Siamese networks \cite{bertinetto2016fully, li2018high, chen2020siamese, tang2022ranking, fu2021onboard, cao2021siamapn++}, have made a remarkable breakthrough in terms of reliability and performance.  
One recent tendency has been to use transformer \cite{vaswani2017attention} models for tracking. CNN is well-suited for localized feature extraction and thus for fine detail and small variation in target detection. Transformers with their attention-based architecture capture interdependencies between different parts in an image and are well-suited for dealing with problems like camera angle variation and scale variation. Hybrid networks inherit advantages from both architectures and thus are best suited for embedded systems and real-world tasks like UAV tracking, where performance and speed are crucial with limited resources.
One successful model in this domain is the Siamese Transformer Pyramid Network (SiamTPN), which combines Siamese architecture with a transformer pyramid network (TPN). This model enhances feature representation and improves tracking performance compared to earlier approaches. However, it still faces certain limitations: it does not utilize temporal information, which reduces the model's robustness in occlusion scenarios and sudden target appearance changes, especially in long-term tracking. Its reliance on cross-correlation performs poorly under severe and complex appearance variations and occlusions, and it delivers suboptimal performance in scenarios with cluttered backgrounds and similar objects. Additionally, it struggles with tracking and precisely localizing small objects.  

In this work, we introduce a novel model called \textbf{T-SiamTPN}, which addresses the limitations of SiamTPN while maintaining computational efficiency and enhancing tracking performance. Large-scale experiments on six aerial tracking benchmarks, as illustrated in Figure~\ref{fig:1}, demonstrate that T-SiamTPN achieves a favorable balance between speed and accuracy. Unlike existing temporal models that suffer from significant speed degradation and increased complexity as the number of templates grows, our method introduces a lightweight and intelligent temporal mechanism that effectively captures inter-frame dependencies with minimal computational overhead. The proposed architecture incorporates one static template and multiple dynamic templates, where dynamic templates are updated selectively and only when the tracker exhibits high confidence. This strategy helps prevent noise accumulation and mitigates model drift due to erroneous updates.

Our model achieves several key improvements:
\begin{enumerate}
    \item It leverages temporal information across frames to enhance robustness against occlusion and improve stability in long-term tracking. In addition, a smart template update mechanism mitigates drift caused by abrupt target appearance changes or background noise.
    
    \item It replaces conventional cross-correlation with a multi-scale transformer attention mechanism, enhancing the model’s ability to distinguish targets in cluttered or complex backgrounds.
    
    \item It integrates a channel attention mechanism into the Pooling Attention (PA) block of the base model and introduces a novel Modulated Pooling Attention (MPA) block. These additions improve feature refinement, resulting in better performance and efficiency, particularly when tracking small objects.
\end{enumerate}


\section{Related Works}
\label{sec:Related Works}

Since the advent of deep networks, visual object tracking using deep learning, especially with Siamese networks, has seen remarkable advancements since 2016 \cite{bertinetto2016fully}. More recently, transformers have been introduced in the fields of computer vision and object tracking, drawing on their success in natural language processing \cite{kugarajeevan2023transformers}. Trackers based on CNNs often struggle due to their limitations in capturing global features and their challenges in dealing with issues like occlusion, scale variations, and complex backgrounds. To address these problems, researchers have started to combine the strengths of CNNs and transformers, aiming to extract and integrate more effective features for improved object tracking.

\subsection{CNN-Transformer-Based Trackers for General Object Tracking}
The first tracker based on CNN-Transformer architecture, called Transformer Meets Tracker \cite{wang2021transformer}, has been launched, consisting of two main pipelines: TrSiam and TrDiMP. TrSiam uses cross-correlation for feature matching, akin to SiamFC \cite{bertinetto2016fully}, while TrDiMP leverages discriminative correlation filters (DCF) \cite{bhat2019learning}. Additionally, the STARK \cite{yan2021learning} model enhances object tracking by incorporating the DETR \cite{carion2020end} object detection transformer, which effectively handles the spatial and temporal aspects of the target. Thanks to its straightforward design, STARK achieves impressive performance and reliability in object tracking.
The ToMP model, similar to STARK, introduces two key improvements: using target position and size as a separate feature map and leveraging transformers for position prediction, enhancing tracking performance and stability.
TransT \cite{chen2021transformer} replaces adaptive correlation with a transformer-based approach for fusing features from both the template and the search region, showing better performance than CNN-based methods. Models such as TrTr \cite{zhao2021trtr} and CTT \cite{zhong2022correlation} enhance the correlation of features between the template and the search region by utilizing transformers, effectively addressing issues like background clutter and long-term variations. To improve both speed and performance, CSWinTT \cite{song2022transformer} and AiATrack \cite{gao2022aiatrack} have been introduced. CSWinTT uses lightweight modules to ensure stable tracking in crowded environments, while AiATrack strikes a balance between high speed and accurate target localization. Furthermore, SparseTT minimizes target drift through optimized structures, leading to improved tracking stability.

Despite the impressive performance attained with such models, integrating deep backbones with large transformer blocks is computationally intensive and requires hardware with high performance. Designing lighter and more efficient models is therefore still an issue, specifically for applications like UAV tracking and embedded systems.

\subsection{CNN-Transformer-Based Trackers for Aerial Object Tracking}
CNN-Transformer-based trackers typically sample features from last CNN backbone feature. Other models, such as HiFT \cite{cao2021hift}, SiamTPN \cite{xing2022siamese}, TCTrack \cite{cao2022tctrack}, SGDViT \cite{yao2023sgdvit}, and PRL-Track \cite{fu2024progressive}, have modified their architectures to specialize in aerial object tracking. HiFT samples features from last three CNN layers and utilizes a transformer for space and semantics processing and sharp foreground-background separation. SiamTPN utilizes a pyramidal transformer for better multi-scale feature and space relationship handling to enhance performance and scale invariance and fast motion robustness. TCTrack utilizes Temporally Adaptive Convolution (TAdaConv) \cite{huang2021tada} and Adaptive Temporal Transformer (AT-Trans) in a temporal framework to achieve high performance and efficiency in embedded systems. SGDViT, designed for UAV tracking, utilizes a dynamic vision transformer and an adaptive saliency-guided process to handle scale variations, background suppression, and aspect ratio variations. PRL-Track utilizes CNN and ViT \cite{dosovitskiy2020image} to achieve object representation with performance in occlusion, scale variations, and dynamic backgrounds.

Recently, several CNN-Transformer-based trackers have been specifically developed for UAV object tracking to overcome the limitations of traditional tracking architectures. 
ORTrack \cite{wu2025learning} enhances robustness against occlusion by applying random spatial masking during training, which enables the model to learn occlusion-invariant representations. Additionally, it employs adaptive feature-based knowledge distillation to train a lightweight variant (ORTrack-D), delivering high accuracy while remaining suitable for deployment on UAV platforms with limited computational resources. 
SGLATrack \cite{xue2025similarity} introduces a similarity-guided, layer-adaptive pruning mechanism that dynamically removes redundant transformer layers based on feature similarity. This approach significantly improves efficiency while preserving competitive tracking accuracy in real-time UAV scenarios. 
UNTrack \cite{qin2025must} leverages a unified transformer framework to fuse multispectral, spatial, and temporal features through an asymmetric architecture and background-suppression mechanism. This enhances target discrimination under challenging conditions, such as cluttered scenes and small-object tracking. 
Aba-ViTrack \cite{li2023adaptive} improves tracking speed and robustness by filtering out irrelevant background tokens using a halting-based mechanism within a single-stream transformer, achieving up to 180 FPS without significant loss in accuracy. 
AVTrack \cite{wu2024learning} introduces a view-invariant feature learning scheme and frame-wise adaptive pruning of transformer blocks to improve robustness under diverse camera perspectives. Furthermore, it incorporates a multi-teacher knowledge distillation strategy to train a compact student network that maintains high performance with reduced computational overhead, making it ideal for embedded UAV applications.
These recent advancements reflect a growing trend toward lightweight and robust hybrid trackers that explicitly address UAV-specific challenges such as occlusion, dynamic viewpoints, cluttered backgrounds, and limited onboard resources. In contrast to conventional deep transformer-based trackers with heavy backbones, modern aerial tracking models emphasize lightweight architectures, real-time processing capabilities, and efficient use of computational resources, making them ideal for deployment in UAV systems with stringent resource constraints.

In this work, T-SiamTPN is designed to enhance SiamTPN and address challenges such as occlusion, long-term variations, cluttered backgrounds, and small objects. By integrating temporal information and replacing cross-correlation with an attention mechanism, it improves tracking accuracy and stability while maintaining efficiency with minimal computational overhead, achieving an optimal balance between performance, speed, and stability.


\section{Proposed Method}
\label{sec:Proposed Method}
\begin{figure*}[t!]
    \centering
    \includegraphics[width=1\textwidth]{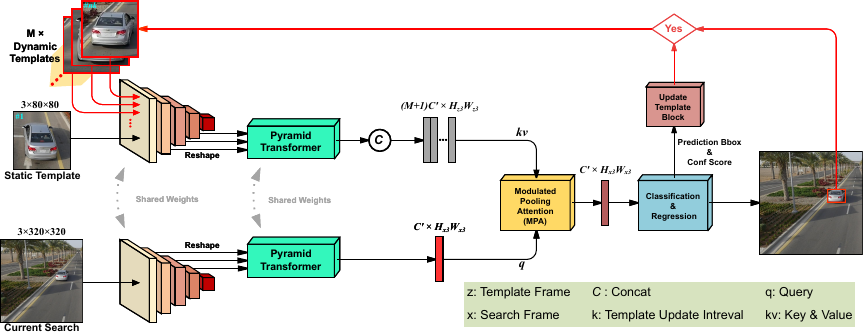}
    \caption{Overview of the T-SiamTPN tracker. Pyramid Transformer (TPN) for feature fusion, a Modulated Pooling Attention (MPA) block for integrating template and search features, for final prediction and Template Update Block for updates best template frame. (Image frames are from UAV123.)}
    \label{fig:2}
\end{figure*}

The T-SiamTPN model, illustrated in Figure~\ref{fig:2}, is an improved SiamTPN with a static template (initial reference) and periodically updated dynamic templates to prevent target loss in difficult cases. Features from both search and template branches in multi-scale are processed with a better pyramid transformer and fused in MPA block. This block produces improved performance in clutter and similar objects via cross-correlation replacement with a transformer-based attention module. Then classification and regression module outputs bounding box and confidence score, and in necessary cases, the template update block updates dynamic templates to attain stable tracking in sequences with a long duration and in cases with occlusion.

\subsection{feature extraction}
The feature extraction network has two pathways: a template branch, which receives crop images from template frames \( z \) with sizes \( 3 \times H_z \times W_z \), and a search branch, which receives the search image from the current frame \( x \) with size \( 3 \times H_x \times W_x \). Both share a shared backbone network for preliminary feature extraction and pyramidal feature map generation (\( P_i \)). Both level 2, 3, and 4 feature maps (\( P_2, P_3, P_4 \)) undergo two steps in TPN block compatibility. Stage 1 processes each feature map using a \( 1 \times 1 \) convolution to downsample from \( C_i \) to \( C' \) channels with different sizes to \( C' \times H_i \times W_i \). Stage 2 processes feature maps \( P'_2, P'_3, P'_4 \) to space-shape flattened vectors with size \( C' \times (H_i W_i) \) to be processed sequentially in the TPN block (\( C' = 192 \)).

\subsection{Improved Pyramidal Transformer Network}
The TPN is given hierarchical feature maps (\( P'_2, P'_3, P'_4 \)) from feature extraction. This multi-scale feature information aggregation network is responsible for generating a stable composite representation and maintaining semantic information in all levels of feature maps. \( P'_2 \) contains low-level information, \( P'_4 \) contains high-level information, and \( P'_3 \) contains mid-level information. \( P'_3 \) is used as query (\( Q \)) in all hierarchical levels to extract inter-level relationships.

\begin{equation}
\begin{aligned}
P_3^{\prime\prime} &= MPAB(P_3^{\prime },P_2^{\prime },P_2^{\prime },R = 4) \\
&+ PAB(P_3^{\prime },P_3^{\prime },P_3^{\prime },R = 2) \\
&+ MPAB(P_3^{\prime },P_4^{\prime },P_4^{\prime },R = 1)
\end{aligned}
\label{eq:tpn_combination}
\end{equation}
\noindent\( P''_3 \) is the combined output of the enhanced feature maps from each stage with dimensions \( P_3^{\prime \prime } \in {R^{C\prime  \times ({H_3}{W_3})}} \).

\noindent The output of the TPN block is a composite feature map with dimensions similar to \( P''_3 \), containing optimized multi-scale information.
\begin{equation}
P_{TPN} = \big\{ PAB(P_3^{\prime \prime },P_3^{\prime \prime },P_3^{\prime \prime },R = 2) \big\}_{n = 2}
\end{equation}

\begin{figure}[t] 
    \centering
    \includegraphics[width=\columnwidth]{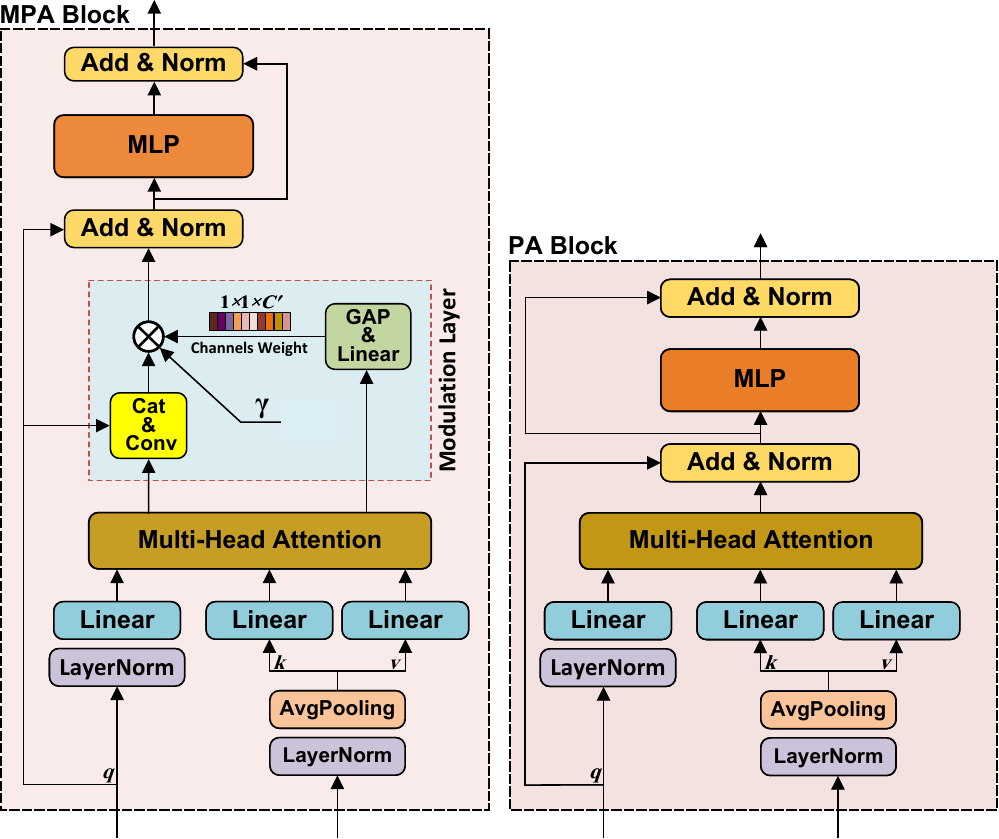} 
    \caption{Comparison between the Pooling Attention (PA) block from the baseline model \cite{xing2022siamese} (right) and the proposed Modulated Pooling Attention (MPA) block (left).}
    \vspace{-5pt}
    \label{fig:3}
\end{figure}

\noindent Figure~\ref{fig:3} illustrates the improved Pyramidal Transformer Network (TPN) with the MPA block.  
The MPA block is explained in Section~\ref{sec:3.3}.

\subsection{Modulated Pooling Attention (MPA)}
\label{sec:3.3}

\noindent As shown in Figure~\ref{fig:3}, MPA is an enhanced version of PA with a query channel attention modulation. This is complemented with space attention for semantic dependency extraction and feature refinement in multi-head attention. MPA's multi-head attention is similar to PA with dependency between \(Q, K,\) and \(V\) being defined as follows:
\vspace{-10pt}  

\begin{equation}
\text{Attention}(Q, K, V) = \text{Softmax} \left(\frac{QK^T}{\sqrt{C}}\right)V
\end{equation}

\noindent The modulation layer is inspired from Hift and TCTrack and strengthens semantic correlations between x and y. Raw feature is x and multi-head attention output is y to boost x. x and y are initially channel-wise concatenated and a 1×1 convolution is then utilized to compress and boost semantic correlation:
\vspace{-15pt} 

\begin{equation}
z = \text{Conv}_{(1 \times 1)}(\text{Concat}(x, y)), \quad z \in \mathbb{R}^{C' \times H_3 \times W_3}
\end{equation}

\noindent The features of \( y \) are first compressed using adaptive average pooling and then passed through two fully connected layers to compute the dynamic weight \( w \) for adjusting \( x \):
\vspace{-10pt} 

\begin{equation}
w = \text{Linear}(\text{AvgPool}(y)), \quad w \in \mathbb{R}^{C' \times 1 \times 1}
\end{equation}

\noindent Finally, \( w \) is applied to \( z \), and the final output is computed as:
\vspace{-15pt} 

\begin{equation}
\text{Output} = x + \gamma \cdot (z \cdot w), \quad \text{Output} \in \mathbb{R}^{C' \times H_3 \times W_3}
\end{equation}

\noindent where \( \gamma \) is a learnable parameter controlling the modulation strength.

\begin{figure}[t] 
    \centering
    \includegraphics[width=\columnwidth]{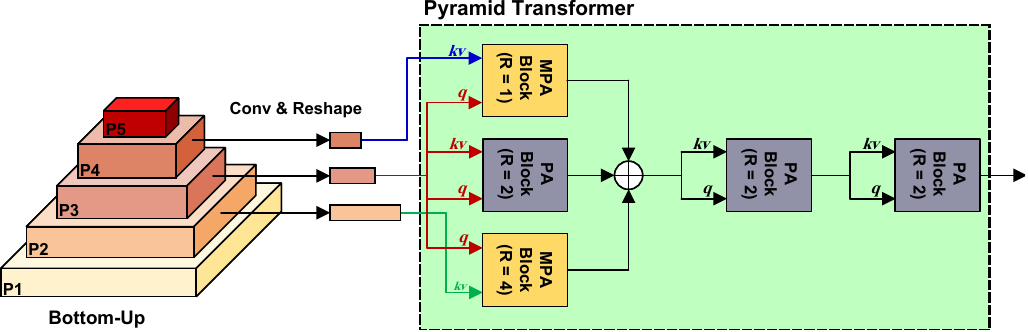} 
    \caption{The enhanced Pyramid Transformer Network (TPN) consists of an MPA block for  cross-stage features integration.}
    \label{fig:4}
\end{figure}

The MPA block imposes a modulation layer to enable semantic interactions between feature maps to capture fine-grained interactions between stages. This is most helpful in cross-attention employed in stages to better extract inter-stage interactions. This block is utilized in Figure~\ref{fig:4}'s feature fusion module to incorporate features from multiple scales in the backbone network.

\subsection{Temporal Modeling in T-SiamTPN}

The T-SiamTPN model is designed specifically to improve tracking performance in difficult cases using a combination of static and dynamic templates. Static template is learned from the first frame and serves as a reliable source for maintaining the original information of the target. Meanwhile, four dynamic templates are updated occasionally to capture target appearance variations, which is crucial in handling occlusion, scale changes, and appearance transformation. As illustrated in Figure~\ref{fig:2}, feature maps from static and dynamic templates are concatenated along the channel dimension to form a multi-channel feature representation and then processed along with feature maps from the search branch in the MPA block.

\begin{figure}[t] 
    \centering
    \includegraphics[width=\columnwidth]{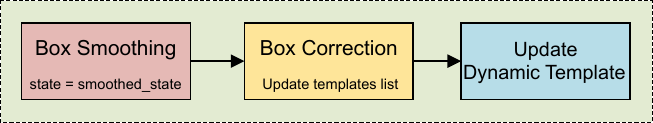} 
    \caption{The execution order of the algorithms used for dynamic template management.}
    \label{fig:5}
\end{figure}

The use is amplified and made more effective with dynamic templates. Incorrect updating can lead to lowered tracking accuracy or losing track. This is balanced with a clever updating process for templates and post-processing algorithms.
Overall, three algorithms are utilized, as shown in Figure~\ref{fig:5}, and which is also indicative of their sequence for execution. They include box smoothing, box correction, and update dynamic template. They are executed in a sequence to provide an integrated workflow to process high accuracy templates optimally.

\subsection{Box Smoothing}

The \textbf{box smoothing algorithm} is designed to reduce sudden and unnatural fluctuations in the position of the bounding box, thereby enhancing the stability and consistency of the tracking process. This method is particularly useful in scenarios where the target is moving rapidly or where previous predictions have exhibited unreasonable variations. By blending the current prediction with the smoothed historical state, the model can provide more reliable and steady tracking.

As shown in \textbf{Algorithm~\ref{alg:box_smoothing}}, the current bounding box ($y_i$) is combined with the previous smoothed state ($S_{\text{prev}}$) to produce a smoothed bounding box ($y_{is}$). This combination is done using a linear weighted average controlled by a smoothing factor $\alpha$, as defined below:

\begin{equation}
y_{is} = \alpha \cdot y_i + (1 - \alpha) \cdot S_{\text{prev}}
\end{equation}

Here, $\alpha$ determines the influence of the current prediction versus the previous state. A higher $\alpha$ gives more weight to the current prediction, making the tracking more responsive but potentially noisier. Conversely, a smaller $\alpha$ results in a smoother output that reacts more slowly to sudden changes.
\vspace{10pt}



\SetAlCapSkip{0.3em}        
\SetAlgoSkip{}              
\SetAlgoInsideSkip{smallskip} 
\SetVlineSkip{0pt}          
\setlength{\algomargin}{1em} 

\begin{algorithm}[H]
\caption{Box Smoothing}\label{alg:box_smoothing}
\KwIn{Current box $y_i$, previous box $S_{\text{prev}}$, smoothing factor $\alpha$}
\KwOut{Smoothed box $y_{is}$}

\If{$\text{Enable\_Smoothing}$ \textbf{and} $S_{\text{prev}}$ exists}{
    $y_{is} \leftarrow \alpha\,y_i + (1-\alpha)\,S_{\text{prev}}$\;
}
\Else{
    $y_{is} \leftarrow y_i$\;
}
\Return $y_{is}$\;
\end{algorithm}

\subsubsection{Box Correction} \label{sec:box_correction}

The primary goal of the \textbf{box‑correction algorithm} is to ensure the stability of the bounding box and to suppress sudden, unrealistic changes in its position or size.  
This procedure becomes crucial whenever predictions exhibit abrupt jumps or the box noticeably drifts from a plausible state.  
As illustrated in \textbf{Algorithm~\ref{alg:box_correction}}, the current smoothed box~($y_{is}$) is compared with the previous box~($S_{\text{prev}}$) to compute two metrics:

\begin{equation}
d = || c_{y_{is}} - c_{S_{\text{prev}}} ||
\end{equation}

\begin{equation}
r = \frac{\text{Area}(y_{is})}{\text{Area}(S_{\text{prev}})}
\end{equation}

\vspace{5pt}

Here, \(d\) denotes the Euclidean distance between the box centres
\(c_{y_{is}}\) and \(c_{S_{\text{prev}}}\),
while \(r\) represents the size ratio.
If either metric exceeds the preset thresholds
\(\theta_d\) (for displacement) or \(\theta_s\) (for scale),
the previous \emph{stable} box is reinstated; otherwise, the current box remains valid.
This safeguard prevents unreasonable deviations and keeps the tracker both stable and accurate.
\vspace{10pt}




    
    
    


\begin{algorithm}[H]
\caption{Box Correction}\label{alg:box_correction}

\KwIn{Current smoothed box $y_{is}$; 
      Previous box $S_{\text{prev}}$; 
      Last stable box $S_s$; 
      Displacement threshold factor $\theta_d$; 
      Size-ratio threshold $\theta_s$}

\KwOut{Corrected box $y_{ic}$; 
       Template list $T = [T_1, T_2, \dots, T_5]$}

\uIf(\tcp*[f]{correction enabled \& previous box available})%
    {$\text{Enable\_Correction}$ \textbf{and} $S_{\text{prev}} \neq \varnothing$}{
    
    $d \leftarrow \text{distance}(c_{y_{is}},\,c_{S_{\text{prev}}})$\;
    $r \leftarrow \dfrac{\operatorname{Area}(y_{is})}{\operatorname{Area}(S_{\text{prev}})}$\;
    $\hat{\theta}_d \leftarrow \theta_d \times \text{diag}(S_{\text{prev}})$\;
    
    \uIf(\tcp*[f]{centre/size deviation too large})%
        {$d > \hat{\theta}_d$ \textbf{or} $r > \theta_s$}{
        
        \uIf{$S_s \neq \varnothing$}{
            $y_{ic} \leftarrow S_s$\;
            $T_{2{:}5} \leftarrow S_s$\;
        }\Else{
            $y_{ic} \leftarrow S_{\text{prev}}$\;
            $T_{2{:}5} \leftarrow S_{\text{prev}}$\;
        }
    }\Else{
        $y_{ic} \leftarrow y_{is}$\;
        $S_s \leftarrow y_{is}$\;
    }
}\Else{
    $y_{ic} \leftarrow y_{is}$\;
}

\KwRet $y_{ic},\,T$\;
\end{algorithm}

\subsubsection{Update Template Dynamic}

This algorithm enhances model suppleness with continuous dynamic updates to the template. Updating is conducted when the current frame number \( i \) is within the interval for updates (\( U \)) and the confidence measure (\( \text{Conf} \)) is larger or equal to the threshold (\( \text{Conf}_{\text{th}} \)), and is given as:

\begin{equation}
i \mod U == 0 \quad \text{and} \quad \text{Conf} > \text{Conf}_{\text{th}}
\end{equation}

\noindent When triggered, the oldest template \( T_2 \) is removed, and the remaining templates shift forward, following:

\begin{equation}
T_2 \leftarrow T_3, \quad T_3 \leftarrow T_4, \quad T_4 \leftarrow T_5, \quad T_5 \leftarrow T_{\text{new}}
\end{equation}

\noindent The static template \( T_1 \) remains unchanged, serving as a stable reference. This efficient update strategy ensures dynamic templates remain relevant, enhancing tracking performance and robustness. For a detailed procedure see \textbf{Algorithm~\ref{alg:template_update}}.
\vspace{10pt}





    
    
    
    


\begin{algorithm}[H]
\caption{Template Update}\label{alg:template_update}

\KwIn{Current frame $x_i$; 
      Current bounding box $y_{ic}$; 
      Template list $T = [T_1, T_2, T_3, T_4, T_5]$; 
      Frame index $i$; 
      Update interval $U$; 
      Confidence threshold $\text{Conf}_{\mathrm{th}}$}

\KwOut{Updated template list $T_u$}

\If{$i \bmod U = 0$ \textbf{and} $\text{Conf}(y_{ic}) \ge \text{Conf}_{\mathrm{th}}$}{
    $T_{\text{new}} \leftarrow \text{ExtractTemplate}(x_i,\,y_{ic})$\;
    $T_2 \leftarrow T_3$\;
    $T_3 \leftarrow T_4$\;
    $T_4 \leftarrow T_5$\;
    $T_5 \leftarrow T_{\text{new}}$\;
    $T_u \leftarrow [T_1, T_2, T_3, T_4, T_5]$\;
}\Else{
    $T_u \leftarrow T$\;
}

\KwRet $T_u$\;
\end{algorithm}

\subsection{Feature Fusion with Attention Mechanism}

In the T-SiamTPN model, the attention mechanism, particularly the MPA block, replaces the cross-correlation method used in the original SiamTPN model to merge information from the search and template branches. As shown in Figure 3-3, the combined feature map from the template branch serves as both the key (\( K \)) and value (\( V \)), while the feature map from the search branch acts as the query (\( Q \)). The dimensions of the template feature map are \( (M+1)C' \times H_{z3} \times W_{z3} \), which includes four dynamic templates (\( M=4 \)) and one static template. In contrast, the search branch feature map has dimensions of \( C' \times H_{x3} \times W_{x3} \).
The attention mechanism effectively models complex and non-linear interactions, offering notable advantages over cross-correlation. While cross-correlation only captures local and linear dependencies, multi-head attention takes into account long-range dependencies, adaptively emphasizing important features and minimizing irrelevant information.

\subsection{Prediction Head}
The prediction head in T-SiamTPN is similar to that in the basic model and consists of classification and regression. The model's goal function is a total of three loss terms:

\begin{equation}
L = \lambda_{cls} L_{cls} + \lambda_{iou} L_{iou} + \lambda_{reg} L_{reg} \end{equation}

\noindent where \( L_{cls} \) is classification cross-entropy loss, \( L_{iou} \) is GIoU loss \cite{rezatofighi2019generalized} for bounding box overlap, and \( L_{reg} \) is L1 loss for coordinate refinement. All the above losses are weighted with \( \lambda \) to enhance tracking performance.


\section{Experiments}

\subsection{Implementation Details}
\textbf{Model:} Similar to the base model, \textit{ShuffleNet\_v2} \cite{zhang2018shufflenet} is used as the backbone, providing outputs from stages 2, 3, and 4, which are compressed to a common dimension of \( C' = 192 \). This model is a pre-trained model on ImageNet dataset. Additionally, pyramidal transformer network has 2 iterations and uses 6 attention heads in each PA and MPA module.

\noindent\textbf{Training:} The model is trained using the LaSOT \cite{fan2019lasot}, GOT-10k \cite{huang2019got}, COCO \cite{lin2014microsoft}, and TrackingNet \cite{muller2018trackingnet} datasets. The search image size is set to \( 320 \times 320 \) pixels, while the template size is configured to \( 80 \times 80 \) pixels. For training, 60,000 samples are utilized, with 10,000 reserved for validation, across 300 epochs. The initial learning rate starts at \( 10^{-4} \) and is decreased to \( 10^{-5} \) at epoch 180, with a decay factor of 0.1. A batch size of 64 is used, and the model is optimized with the ADAMW 
\cite{loshchilov2017decoupled} algorithm. The loss function is weighted as follows:
\[
\lambda_{cls} = 5, \quad \lambda_{iou} = 2, \quad \lambda_{reg} = 2
\]

\noindent\textbf{Inference:} The model is tested with an Nvidia RTX 3090 GPU and run on Nvidia Jetson Nano edge hardware.
For tracking performance evaluation, the one-pass evaluation (OPE) metrics are used, including precision, normalized precision, and success rate. These metrics are expressed as percentages (\%).

\subsection{Performance Comparison}
The performance is compared with other latest models for aerial tracking, i.e., SiamTPN \cite{xing2022siamese}, HIFT \cite{cao2021hift}, SiamAPN \cite{fu2021onboard}, SiamAPN++ \cite{cao2021siamapn++}, LPAT \cite{fu2022local}, FDNT \cite{zuo2022end}, Siam Trans \cite{sun2022siamese}, SiamSA \cite{zheng2022siamese}, UDAT\_CAR \cite{ye2022unsupervised}, UDAT\_BAN \cite{ye2022unsupervised}, TCTrack \cite{cao2022tctrack}, TCTrack++ \cite{cao2023towards}, SGDViT \cite{yao2023sgdvit}, PRL-Track \cite{fu2024progressive}, and TFITrack \cite{hu2024tfitrack}. All the trackers are executed entirely with their public codes for a fairer comparison. Though results are slightly different from the original publications, the same settings and evaluation measures are adopted for all models.

In selecting baseline models for comparison, three main criteria were considered: (1) availability of official code and pretrained weights, (2) relevance to the UAV tracking domain, and (3) architectural similarity (preferably CNN–Transformer hybrids). Many recent methods were excluded because they lacked complete code or pretrained weights at the time of writing, making direct and fair comparison infeasible.

\noindent In the comparison tables showing performance results, color coding is employed to highlight the top ranks. Specifically, \textbf{\textcolor{red}{red}}, \textbf{\textcolor{green}{green}}, and \textbf{\textcolor{blue}{blue}} indicate first, second, and third places, respectively.

\subsubsection{Performance Comparison on LaSOT and GOT-10k Test Sets}

Trackers' performance on the LaSOT dataset is measured by success rate, precision, and normalized precision. On the GOT-10k dataset, performance is measured on success rates at threshold 0.5 and 0.75 and Average Overlap (AO) metric. The results are reported in Table ~\ref{tab:1}.

\begin{table}[b!] 
    \centering
    \renewcommand{\arraystretch}{1} 
    \setlength{\tabcolsep}{8pt} 
    \resizebox{\linewidth}{!}{ 
    \begin{tabular}{@{}>{\centering\arraybackslash}m{4cm} | ccc | ccc@{}} 
        \toprule
        \multirow{2}{*}{\shortstack{\textbf{Tracker} \\ \textbf{Method}}} & \multicolumn{3}{c|}{\textbf{LaSOT}} & \multicolumn{3}{c}{\textbf{GOT-10k}} \\ 
        & \textbf{Succ.} & \textbf{Prec$_{\text{norm.}}$} & \textbf{Prec.} & \textbf{AO} & \textbf{SR$_{0.5}$} & \textbf{SR$_{0.75}$} \\
        \midrule
        SiamTPN \cite{xing2022siamese}     & \textcolor{green}{\textbf{53.54}} & \textcolor{blue}{\textbf{61.72}} & \textcolor{green}{\textbf{51.41}} & \textcolor{green}{\textbf{55.2}} & \textcolor{green}{\textbf{64.5}} & \textcolor{blue}{\textbf{44.0}} \\
        HIFT \cite{cao2021hift}        & 45.1  & 52.7  & 42.1  & 49.4 & 58.6 & 27.5 \\
        SiamAPN \cite{fu2021onboard}     & 43.5  & 51.2  & 44.2  & 44.0 & 48.0 & 23.5 \\
        SiamAPN++ \cite{cao2021siamapn++}   & 42.4  & 50.3  & 43.3  & 47.6 & 54.5 & 27.4 \\
        LPAT \cite{fu2022local}        & 44.5  & 51.4  & 42.7  & 50.1 & 57.9 & 31.2 \\
        FDNT \cite{zuo2022end}        & 49.0  & 55.2  & \textcolor{blue}{\textbf{47.1}}  & \textcolor{blue}{\textbf{53.6}} & \textcolor{blue}{\textbf{61.6}} & 36.7 \\
        Siam Trans \cite{sun2022siamese}  & -     & 50.5  & 43.6  & 53.2 & 60.9 & 41.3 \\
        SiamSA \cite{zheng2022siamese}      & 47.0  & 54.4  & 46.9  & 50.4 & 59.1 & 31.7 \\
        UDAT\_CAR \cite{ye2022unsupervised}   & 42.1  & 49.9  & 42.4  & 39.5 & 42.5 & 14.7 \\
        UDAT\_BAN \cite{ye2022unsupervised}   & 46.7  & 57.1  & 48.0  & 42.6 & 47.4 & 17.4 \\
        TCTrack \cite{cao2022tctrack}     & 46.2  & 53.9  & 46.9  & 48.4 & 55.9 & 31.1 \\
        TCTrack++ \cite{cao2023towards}   & 46.5  & 54.1  & 46.9  & 52.9 & 61.3 & 35.8 \\
        SGDViT \cite{yao2023sgdvit}      & 46.2  & 53.8  & 45.7  & 51.3 & 59.9 & 33.0 \\
        PRL-Track \cite{fu2024progressive}   & 46.6  & 53.7  & 45.4  & 50.9 & 59.4 & 30.4 \\
        TFITrack \cite{hu2024tfitrack}    & \textcolor{blue}{\textbf{52.6}}  & \textcolor{blue}{\textbf{62.2}}  & 30.4  & 49.1 & -    & \textcolor{red}{\textbf{57.0}} \\
        \rowcolor{gray!20}
        \textbf{T-SiamTPN (ours)} & \textcolor{red}{\textbf{56.64}} & \textcolor{red}{\textbf{65.52}} & \textcolor{red}{\textbf{56.68}} & \textcolor{red}{\textbf{58.0}} & \textcolor{red}{\textbf{68.2}} & \textcolor{green}{\textbf{45.9}} \\
        \bottomrule
    \end{tabular}
    } 
    \caption{Performance comparison of T-SiamTPN with state-of-the-art aerial trackers on LaSOT and GOT-10k.}
    \label{tab:1}
\end{table}

\subsubsection{Performance Comparison on UAV123, UAV123@10fps and UAV20L}

\begin{figure*}[t!]
    \centering
    \includegraphics[width=1\textwidth]{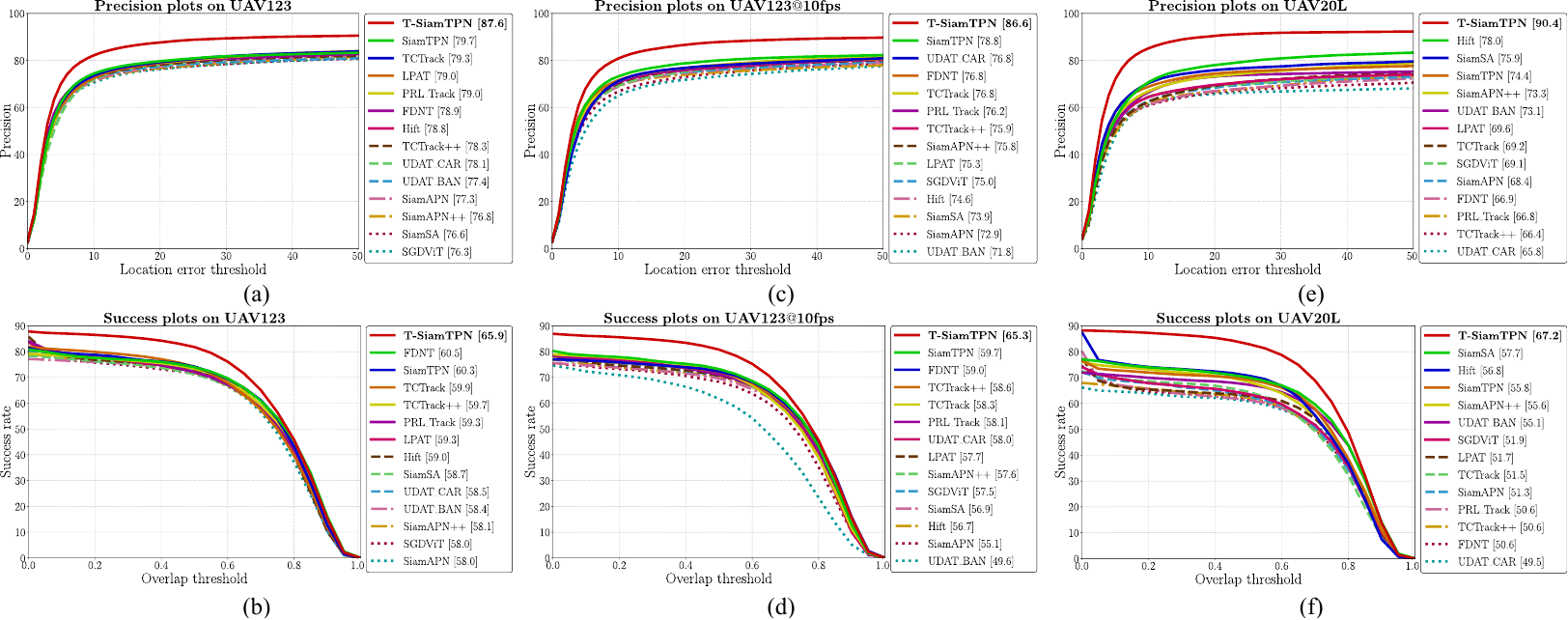}
    \caption{Performance comparison of T-SiamTPN with state-of-the-art aerial trackers on UAV123, UAV123@10fps, and UAV20L benchmarks, presented through success rate and precision curves for (a, b) UAV123, (c, d) UAV123@10fps, and (e, f) UAV20L datasets.}
    
    \label{fig:6}
\end{figure*}

Trackers' performance on UAV123, UAV123@10fps and UAV20L datasets is measured as success rate and precision at various threshold values. The performance is shown in Figure \ref{fig:6}.

\noindent\textbf{UAV123 \cite{benchmark2016benchmark}:} The UAV123 dataset contains 123 aerial video sequences comprising more than 110,000 frames, offering a benchmark to test the performance of trackers under various scenarios.
It comprises 115 sequences of real-world drones and 8 sequences of a drone simulator, split based on 12 tracking difficulties such as occlusion, scale change, rotation, illumination change, and fast motion. As seen from Figures \ref{fig:6}-a and \ref{fig:6}-b, T-SiamTPN surpassed all trackers with precision of 87.6\% and success rate of 65.9\%.

\noindent\textbf{UAV123@10fps \cite{benchmark2016benchmark}:} This dataset is derived from the original UAV123 dataset, downsampled to 10 frames per second, and contains over 35,000 frames. The lower frame rate increases the temporal gap between consecutive frames, adding further challenges to object tracking. As shown in Figures \ref{fig:6}-c and \ref{fig:6}-d, T-SiamTPN again achieved the best performance among all trackers, with a precision of 86.6\% and a success rate of 65.3\%.

\noindent\textbf{UAV20L \cite{benchmark2016benchmark}:} It is a subset of UAV123, which consists of 20 longer video sequences with over 55,000 frames. It is a special set for aerial tracking evaluation in long-term scenarios. As observed from Figures \ref{fig:6}-e and \ref{fig:6}-f, T-SiamTPN achieved the best among all trackers with precision of 90.4\% and a success rate of 67.21\%.

\subsubsection{Performance Comparison on UAVDT, UAVTrack112 and UAVTrack112L}

Trackers' performance on the UAVDT, UAVTrack112, and UAVTrack112L datasets is evaluated by comparing their success rate and precision. Results are shown in Table \ref{tab:2}.
\noindent\textbf{UAVDT \cite{du2018unmanned}:} This dataset has 112 real-world drone experiment video sequences. It has more than 100,000 frames and 13 challenging attributes like fast motion, low resolution, long-term tracking, and aspect ratio changes that make target tracking more complex.
As shown in Table \ref{tab:2}, T-SiamTPN showed competitive results with an precision of 82.4\% and a success rate of 62.79\%.

\noindent\textbf{UAVTrack112 \cite{cao2021siamapn++}:} This dataset consists of 112 video sequences recorded during real-world drone experiments. It contains over 100,000 frames and includes 13 challenging attributes such as fast motion, low resolution, long-term tracking, and aspect ratio changes, making target tracking more difficult.
As shown in Table \ref{tab:2}, T-SiamTPN demonstrated competitive performance, achieving a precision of 82.4\% and a success rate of 62.79\%.

\noindent\textbf{UAVTrack112 \cite{cao2021siamapn++}:}It is a subset of UAVTrack112 for long-term aerial tracking. It consists of 45 video sequences with around 70,000 frames in total. As can be seen in Table \ref{tab:2}, T-SiamTPN is competitive with 82.3\% precision and a success rate of 60.71\%..

\begin{table}[t!]
    \centering
    \renewcommand{\arraystretch}{1}
    \setlength{\tabcolsep}{8pt}
    \resizebox{\linewidth}{!}{
    \begin{tabular}{@{}>{\centering\arraybackslash}m{4cm} | c c | c c | c c @{}}
        \toprule
        \multirow{2}{*}{\shortstack{\textbf{Tracker} \\ \textbf{Method}}} & \multicolumn{2}{c|}{\textbf{UAVDT}} & \multicolumn{2}{c|}{\textbf{UAVTrack112}} & \multicolumn{2}{c}{\textbf{UAVTrack112L}} \\
        & \textbf{Succ.} & \textbf{Prec.} & \textbf{Succ.} & \textbf{Prec.} & \textbf{Succ.} & \textbf{Prec.} \\
        \midrule
        SiamTPN \cite{xing2022siamese}     & 43.8  & 56.8  & 58.9  & 76.3  & 53.33 & 72.1  \\
        HIFT \cite{cao2021hift}            & 46.8  & 63.8  & 56.0  & 72.7  & 53.1  & 70.8  \\
        SiamAPN \cite{fu2021onboard}       & 51.9  & 71.0  & 61.4  & \textcolor{blue}{\textbf{81.1}}  & 57.8  & 78.0  \\
        SiamAPN++ \cite{cao2021siamapn++}  & 54.4  & \textcolor{green}{\textbf{74.8}}  & 58.8  & 77.3  & 53.7  & 73.1  \\
        LPAT \cite{fu2022local}            & 53.8  & 71.6  & 60.8  & 77.9  & 57.0  & 76.3  \\
        FDNT \cite{zuo2022end}             & \textcolor{green}{\textbf{55.0}}  & 72.8  & 60.8  & 77.5  & 58.7  & 77.3  \\
        SiamSA \cite{zheng2022siamese}     & \textcolor{blue}{\textbf{54.7}}  & \textcolor{red}{\textbf{75.0}}  & 61.2  & 80.0  & 56.8  & 76.4  \\
        UDAT\_CAR \cite{ye2022unsupervised} & 46.1  & 69.3  & 53.0  & 72.0  & 53.7  & 72.28 \\
        UDAT\_BAN \cite{ye2022unsupervised} & 49.2  & 72.7  & 57.5  & 79.1  & 55.9  & 76.0  \\
        TCTrack \cite{cao2022tctrack}      & 52.2  & 70.0  & 58.8  & 75.8  & 53.7  & 72.0  \\
        TCTrack++ \cite{cao2023towards}    & 52.6  & 72.4  & \textcolor{blue}{\textbf{62.2}}  & 79.3  & \textcolor{blue}{\textbf{60.0}}  & \textcolor{blue}{\textbf{78.7}}  \\
        SGDViT \cite{yao2023sgdvit}        & 49.6  & 65.5  & 59.0  & 75.8  & 55.1  & 73.8  \\
        PRL-Track \cite{fu2024progressive} & 54.0  & 71.4  & \textcolor{green}{\textbf{62.7}}  & \textcolor{green}{\textbf{81.7}}  & \textcolor{red}{\textbf{61.0}}  & \textcolor{red}{\textbf{83.1}}  \\
        \rowcolor{gray!20}
        \textbf{T-SiamTPN (ours)} & \textcolor{red}{\textbf{55.5}} & \textcolor{blue}{\textbf{73.41}} & \textcolor{red}{\textbf{62.79}} & \textcolor{red}{\textbf{82.4}} & \textcolor{green}{\textbf{60.71}} & \textcolor{green}{\textbf{82.3}} \\
        \bottomrule
    \end{tabular}
    }
    \caption{Performance comparison of T-SiamTPN with state-of-the-art aerial trackers on UAVDT, UAVTrack112, and UAVTrack112L benchmarks.}
    \label{tab:2}
\end{table}

\subsection{Attribute-based Comparison}
The performance of T-SiamTPN in tackling complex aerial tracking challenges is assessed and compared through feature-based analysis on the UAV123 dataset. The results are shown in Table \ref{tab:3}. Findings reveal that T-SiamTPN surpasses all other models in every scenario. Notably, in cases of partial occlusion (POC) and full occlusion (FOC), the model effectively utilizes its temporal capabilities to recover the target after occlusion, showcasing a robust spatial-temporal understanding. Furthermore, in situations involving background clutter (BC) and the presence of similar objects (SOB), the attention mechanism in the feature fusion between the template and search branches allows the model to better grasp global relationships, enabling it to accurately identify the target without confusion.

\renewcommand{\arraystretch}{1}
\setlength{\tabcolsep}{4pt}

\begin{table}[b!]
    \centering
    \resizebox{\columnwidth}{!}{
    \begin{tabular}{>{\centering\arraybackslash}m{4cm} | 
                    >{\centering\arraybackslash}m{1cm} 
                    >{\centering\arraybackslash}m{1cm} | 
                    >{\centering\arraybackslash}m{1cm} 
                    >{\centering\arraybackslash}m{1cm} | 
                    >{\centering\arraybackslash}m{1cm} 
                    >{\centering\arraybackslash}m{1cm} | 
                    >{\centering\arraybackslash}m{1cm} 
                    >{\centering\arraybackslash}m{1cm} | 
                    >{\centering\arraybackslash}m{1cm} 
                    >{\centering\arraybackslash}m{1cm} }  
    \toprule
   \multirow{2}{*}{\shortstack{\textbf{Tracker} \\ \textbf{Method}}}  
    & \multicolumn{2}{c}{\textbf{BC}} 
    & \multicolumn{2}{c}{\textbf{FOC}}
    & \multicolumn{2}{c}{\textbf{LR}}
    & \multicolumn{2}{c}{\textbf{POC}}
    & \multicolumn{2}{c}{\textbf{SOB}} \\
    
    & \textbf{Succ.} & \textbf{Prec.}
    & \textbf{Succ.} & \textbf{Prec.}
    & \textbf{Succ.} & \textbf{Prec.}
    & \textbf{Succ.} & \textbf{Prec.}
    & \textbf{Succ.} & \textbf{Prec.}\\
    \midrule

    SiamTPN \cite{xing2022siamese}
    & \textbf{\textcolor{green}{43.1}} & \textbf{\textcolor{blue}{63.4}}
    & \textbf{\textcolor{green}{41.4}} & \textbf{\textcolor{green}{65.0}}
    & 43.2 & 64.7
    & \textbf{\textcolor{green}{52.1}} & \textbf{\textcolor{green}{71.8}}
    & \textbf{\textcolor{green}{53.4}} & \textbf{\textcolor{blue}{74.0}} \\

    HIFT \cite{cao2021hift}
    & 38.6 & 58.6
    & 35.4 & 58.3
    & 42.5 & 65.1
    & 48.9 & 68.6
    & 51.1 & 71.3 \\

    SiamAPN \cite{fu2021onboard}
    & 40.0 & 60.7
    & 32.3 & 55.0
    & 41.8 & 64.8
    & 47.5 & 66.3
    & 47.9 & 65.8 \\

    SiamAPN++ \cite{cao2021siamapn++}
    & 38.6 & 58.2
    & 34.5 & 57.5
    & 41.4 & 63.2
    & 48.1 & 66.4
    & 49.4 & 67.7 \\

    LPAT \cite{fu2022local}
    & \textbf{\textcolor{blue}{43.0}} & \textbf{\textcolor{green}{64.3}}
    & 36.5 & 62.1
    & 42.3 & 66.5
    & 49.4 & 69.9
    & 50.1 & 70.4 \\

    FDNT \cite{zuo2022end}
    & 40.9 & 59.0
    & 34.8 & 55.3
    & 43.6 & 65.0
    & 50.7 & 69.6
    & 51.2 & 69.2 \\

    SiamSA \cite{zheng2022siamese}
    & 34.9 & 51.7
    & 35.5 & 57.5
    & \textbf{\textcolor{blue}{44.1}} & 66.4
    & 47.2 & 65.2
    & 49.9 & 68.6 \\

    UDAT CAR \cite{ye2022unsupervised}
    & 37.9 & 56.2
    & 30.1 & 49.4
    & \textbf{\textcolor{green}{45.4}} & \textbf{\textcolor{green}{67.1}}
    & 48.3 & 66.9
    & 50.7 & 69.4 \\

    UDAT BAN \cite{ye2022unsupervised}
    & 36.5 & 54.9
    & 32.7 & 53.5
    & 41.0 & 62.1
    & 48.1 & 66.9
    & 48.8 & 65.8 \\

    TCTrack \cite{cao2022tctrack}
    & 41.7 & 61.8
    & 35.9 & 57.7
    & 43.3 & \textbf{\textcolor{blue}{66.9}}
    & \textbf{\textcolor{blue}{50.9}} & \textbf{\textcolor{blue}{70.3}}
    & \textbf{\textcolor{blue}{53.3}} & \textbf{\textcolor{green}{74.4}} \\

    TCTrack++ \cite{cao2023towards}
    & 42.2 & 62.3
    & \textbf{\textcolor{blue}{38.2}} & 60.7
    & 42.8 & 65.2
    & 50.0 & 69.2
    & 52.0 & 70.0 \\

    SGDViT \cite{yao2023sgdvit}
    & 36.5 & 53.9
    & 35.6 & 57.2
    & 40.2 & 61.5
    & 47.5 & 65.6
    & 48.2 & 66.5 \\

    PRL-Track \cite{fu2024progressive}
    & 41.7 & 63.0
    & 37.7 & \textbf{\textcolor{blue}{62.3}}
    & 43.6 & \textbf{\textcolor{green}{67.1}}
    & 50.2 & 70.0
    & 51.3 & 71.0 \\

    \rowcolor{gray!20}
    \textbf{T-SiamTPN (ours)}
    & \textbf{\textcolor{red}{52.7}} 
    & \textbf{\textcolor{red}{77.2}} 
    & \textbf{\textcolor{red}{49.3}} 
    & \textbf{\textcolor{red}{75.6}} 
    & \textbf{\textcolor{red}{50.7}} 
    & \textbf{\textcolor{red}{77.0}} 
    & \textbf{\textcolor{red}{60.3}} 
    & \textbf{\textcolor{red}{83.7}} 
    & \textbf{\textcolor{red}{61.5}} 
    & \textbf{\textcolor{red}{84.5}} \\

    \midrule 
    
    \textbf{$\Delta(\%)$}
    & 22.3$\uparrow$ 
    & 20.0$\uparrow$
    & 19.1$\uparrow$
    & 16.3$\uparrow$
    & 11.7$\uparrow$
    & 14.7$\uparrow$
    & 15.7$\uparrow$
    & 16.6$\uparrow$
    & 15.1$\uparrow$
    & 13.6$\uparrow$ \\

    \bottomrule
    \end{tabular}
    }
    \caption{Attribute-based evaluation of T-SiamTPN and state-of-the-art aerial trackers on the UAV123 benchmark.  
\( \Delta \) denotes the improvement compared to the second-best tracker.}
    \label{tab:3}
\end{table}

\subsection{Performance and Computational Analysis}
The T-SiamTPN is designed to achieve an optimal balance between efficiency and accuracy. With a light-weight framework and temporal functionality, state-of-the-art performance is achieved with minimal added complexity in computational complexity.
Table \ref{tab:4} provides a summary of a detailed comparison between prominent measures, i.e., number of parameters, floating-point operations per second (FLOPs), and inference rate (FPS) on embedded Nvidia Jetson Nano. With 7.33M parameters, 2.017 GFLOPs, and an inference speed of 7.1 FPS, T-SiamTPN stands out as one of the most lightweight trackers.

\begin{table}[t!]
    \centering
    \renewcommand{\arraystretch}{1} 
    \setlength{\tabcolsep}{8pt} 
    \resizebox{\linewidth}{!}{ 
    \begin{tabular}{c|c|c|c|c}  
    
        \toprule
        \textbf{Tracker Method} & \textbf{\#Param} & \textbf{FLOPs} & \textbf{Avg. Speed} & \textbf{Avg. Prec.} \\
        \midrule
        SiamTPN \cite{xing2022siamese}    & \textcolor{red}{\textbf{6.171 M}} & \textcolor{red}{\textbf{1.674 G}} & \textcolor{blue}{\textbf{7.4} \textit{fps}} & 73.01 \\
        HIFT \cite{cao2021hift}       & 11.066 M  & 5.629 G  & 7.3 \textit{fps}  & 73.11 \\
        SiamAPN \cite{fu2021onboard}   & 15.149 M  & 7.62 G   & 6.7 \textit{fps}  & 74.78 \\
        SiamAPN++ \cite{cao2021siamapn++} & 15.374 M  & 7.239 G  & 7 \textit{fps}    & 75.18 \\
        LPAT \cite{fu2022local}       & \textcolor{blue}{\textbf{10.078 M}} & \textcolor{blue}{\textbf{5.572 G}} & 5.8 \textit{fps}  & 74.95 \\
        FDNT \cite{zuo2022end}       & 15.783 M  & 6.548 G  & 6.7 \textit{fps}  & 75.03 \\
        SiamSA \cite{zheng2022siamese}    & 15.49 M   & 13.078 G & 2.9 \textit{fps}  & \textcolor{blue}{\textbf{76.3}}  \\
        UDAT\_CAR \cite{ye2022unsupervised}  & 52.867 M  & 59.307 G & 1.2 \textit{fps}  & 72.38 \\
        UDAT\_BAN \cite{ye2022unsupervised}  & 55.414 M  & 59.537 G & 1.2 \textit{fps}  & 75.02 \\
        TCTrack \cite{cao2022tctrack}   & 10.396 M  & 9.782 G  & \textcolor{green}{\textbf{8.4} \textit{fps}} & 73.7  \\
        TCTrack++ \cite{cao2023towards} & 15.207 M  & 10.012 G & \textcolor{red}{\textbf{8.5} \textit{fps}} & 75.31 \\
        SGDViT \cite{yao2023sgdvit}    & 23.58 M   & 8.432 G  & 5.3 \textit{fps}  & 72.58 \\
        PRL-Track \cite{fu2024progressive}  & 24.86 M   & 5.903 G  & 6.4 \textit{fps}  & \textcolor{green}{\textbf{76.4}}  \\
        \rowcolor{gray!20}
        \textbf{T-SiamTPN (ours)} & \textcolor{green}{\textbf{7.33 M}} & \textcolor{green}{\textbf{1.998 G}} & 7.1 \textit{fps} & \textcolor{red}{\textbf{83.79}} \\
        \bottomrule
    \end{tabular}
    }
    \caption{Comparison of state-of-the-art aerial trackers in terms of parameter count, computational complexity, speed on Jetson Nano, and average precision. The proposed T-SiamTPN achieves the best precision while maintaining efficiency.}
    \label{tab:4}
\end{table}

\subsection{Qualitative Evaluation} 
\begin{figure*}[t!]
    \centering
    \includegraphics[width=1\textwidth]{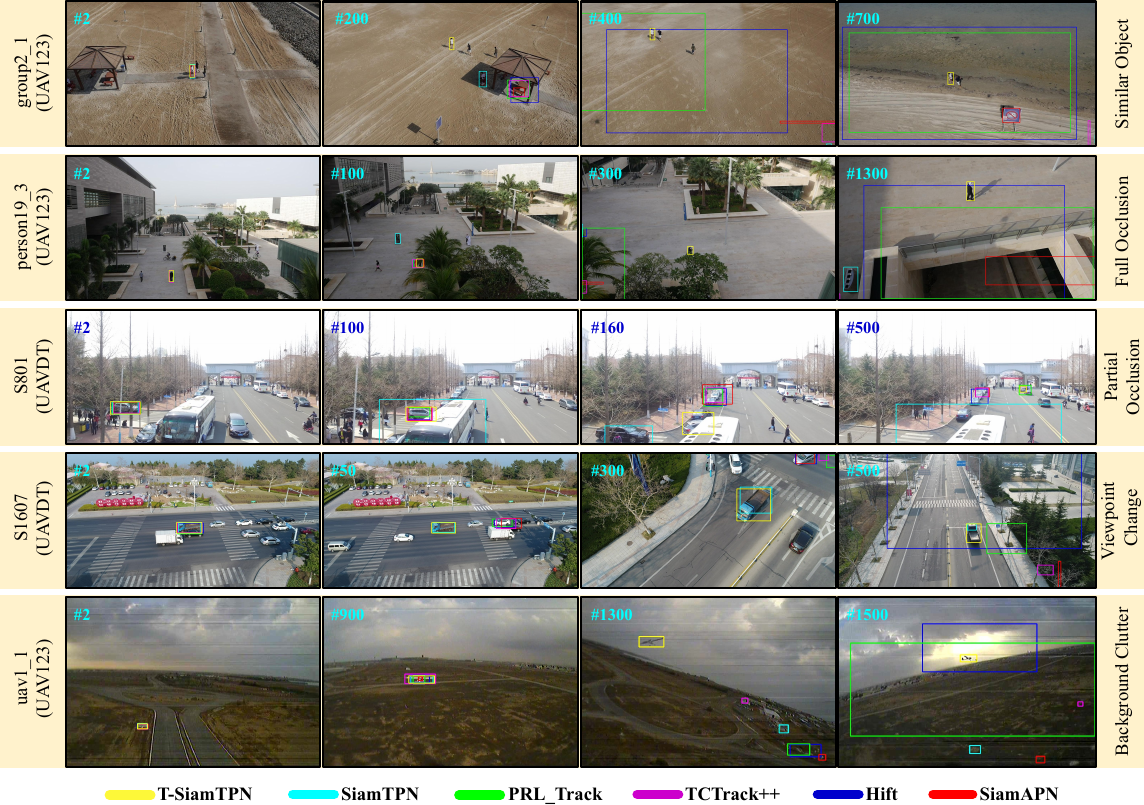}
    \caption{Comparison of T-SiamTPN with state-of-the-art aerial trackers across different scenarios from UAV123 and UAVDT benchmarks, including partial occlusion, viewpoint changes, full occlusion, and background clutter.}
    \label{fig:7}
\end{figure*}

For a qualitative, intuitive visualization of the performance of the tracker, qualitative evaluations are depicted in Figure \ref{fig:7}. The proposed tracker, T-SiamTPN, is compared here with five most advanced trackers like SiamTPN \cite{xing2022siamese}, Hift \cite{cao2021hift}, TCTrack++ \cite{cao2023towards}, SiamAPN \cite{fu2021onboard}, and PRL-Track \cite{fu2024progressive}.
For comparison, five video sequences from the UAV123 \cite{benchmark2016benchmark} and UAVDT \cite{du2018unmanned} benchmarks have been selected, with challenges such as Similar Objects (SOB), Full Occlusion (FOC), Partial Occlusion (POC), Viewpoint Change (VC), and Background Clutter (BC).
Five frames selected to display the output at different time intervals are provided in every video. The leftmost column represents the initial frame of tracking, and the columns to its right represent some selected intermediate frames of the sequence. Bounding boxes predicted by  trackers are represented in each frame. Yellow Bounding boxes represents the proposed model T-SiamTPN, and other colors represent the rest of the trackers.
The performance in all of cases reveals that T-SiamTPN possesses the best performance. For example, in Full Occlusion (FOC) challenge, T-SiamTPN model achieves ideal recovery of the target after occlusion disappearance.

\subsection{Ablation Study}

Ablation studies have been carried out to find contribution from different components to T-SiamTPN performance. This subsection considers different numbers of template frames. Comparison is also given between basic and enhanced variants to find contribution from different components to performance.

\noindent\textbf{Number of Template Frames:}
The results of the experiments on the UAV123 and UAV123@10fps benchmarks, as shown in Table~\ref{tab:5}, suggest that increasing the number of template frames enhances the model’s performance. Notably, utilizing five template frames achieves the highest precision and success rate. The purpose of Table~\ref{tab:5} was not merely to identify the best configuration of our model, but rather to validate the effectiveness of the proposed temporal mechanism. These results indicate that increasing the number of templates up to five, within the context of our smart and adaptive temporal framework, leads to significant performance gains while introducing minimal overhead in terms of speed and computational cost. Please note that our update strategy is an adaptive mechanism, rather than a rigid or fixed one. It is specifically designed to prevent incorrect template updates that could lead to tracker confusion and performance degradation.

\newcolumntype{C}[1]{>{\centering\arraybackslash}p{#1}}

\begin{table}[t]
    \centering
    \renewcommand{\arraystretch}{1} 
    \setlength{\tabcolsep}{4pt} 
    
    \resizebox{1\linewidth}{!}{
    \begin{tabular}{
        C{4cm} | C{2cm} C{2cm} | C{2cm} C{2cm}
    }
        \toprule
       \multirow{2}{*}{\shortstack{\textbf{Template Frame}}}
        & \multicolumn{2}{C{4cm}|}{\textbf{UAV123}} 
        & \multicolumn{2}{C{4cm}}{\textbf{UAV123\char64 10fps}} \\
        & \textbf{Succ.} & \textbf{Prec.} & \textbf{Succ.} & \textbf{Prec.} \\
        \midrule
        \textbf{2} & 62.91 & 83.0 & 62.44 & 82.71 \\
        \textbf{3} & 64.59 & 85.32 & 63.48 & 84.1 \\
        \textbf{4} & 65.52 & 86.18 & 64.73 & 85.15 \\
        \textbf{5} & \textbf{65.94} & \textbf{87.63} & \textbf{65.3} & \textbf{86.59} \\
        \bottomrule
    \end{tabular}
    }
    \caption{Impact of template frames on performance. Using five frames achieves the best success rate and precision on UAV123 and UAV123@10fps.}
    \label{tab:5}
\end{table}

\begin{figure}[t] 
    \centering
    \includegraphics[width=\columnwidth]{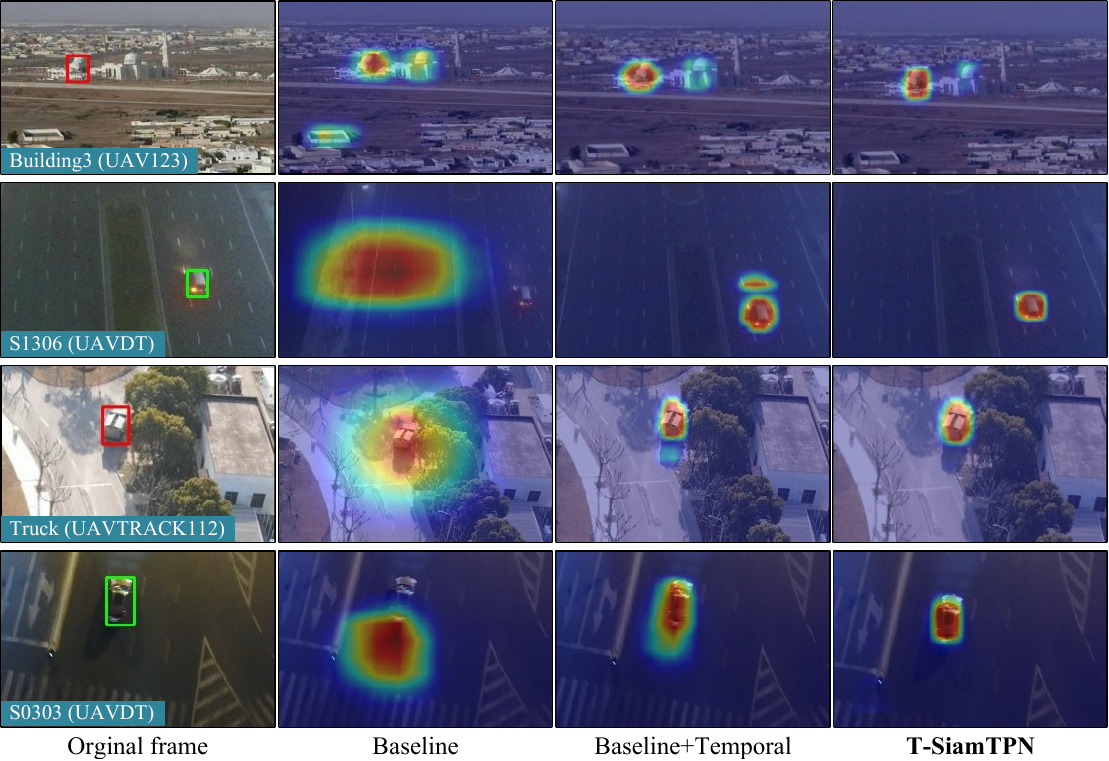} 
    \caption{Qualitative results and heatmaps of the baseline model, temporal-model, and T-SiamTPN on UAV123, UAVDT, and UAVTrack112 datasets. The first column shows the initial frame with the ground truth bounding box.}
    \label{fig:8}
\end{figure}

\noindent\textbf{Effectiveness of Model Components:}
To assess the influence of essential elements in T-SiamTPN, we carried out a series of experiments across all six aerial tracking benchmark datasets. The qualitative findings are illustrated in Figure \ref{fig:8}, and the quantitative results are compiled in Table \ref{tab:6}. The precision and success rate figures in the table reflect the average performance across these six aerial benchmarks.

\begin{table}[t]
    \centering
    \renewcommand{\arraystretch}{1} 
    \setlength{\tabcolsep}{6pt} 
    
    \resizebox{1\linewidth}{!}{
    \begin{tabular}{
        C{5.2cm} | C{1.6cm} C{1.6cm} | C{1.6cm} C{1.6cm}
    }
        \midrule
        \multirow{2}{*}{\textbf{Model}} 
        & \multicolumn{2}{C{3.6cm}|}{\textbf{Succ.}} 
        & \multicolumn{2}{C{3.6cm}}{\textbf{Prec.}} \\
        & \textbf{Value} & $\boldsymbol{\Delta}_{\text{Succ}}$(\%)
 & \textbf{Value} & $\boldsymbol{\Delta}_{\text{Prec}}$(\%) \\
        \midrule
        \textbf{Baseline} & 55.29 & - & 73.01 & - \\
        \textbf{Baseline+Temporal} & 59.26 & +3.97 & 78.62 & +5.61 \\
        \textbf{Baseline+Temporal+AC} & 62.46 & +7.17 & 82.04 & +9.03 \\
        \textbf{\textbf{T-SiamTPN}} & \textbf{62.9} & \textbf{+7.61} & \textbf{83.79} & \textbf{+10.78} \\
        \bottomrule
    \end{tabular}
    }
    \caption{Impact of T-SiamTPN components: AC (attention mechanism combiner) replaces cross-correlation for feature combining. T-SiamTPN consist of the temporal modeling, AC, and MPA block.}

    \label{tab:6}
\end{table}

\noindent\textbf{Templates Features Combination Methods: }
For combining the features of the template branch, three different methods have been examined: concatenation (Concat), averaging (Mean), and summation (Sum). Each of these methods has a different impact on precision, success rate, the number of parameters, and execution speed.
The experimental results shown in Table \ref{tab:7} on the UAV123@10fps \cite{benchmark2016benchmark} benchmark indicate that the concatenation (Concat) method provides the best performance in terms of accuracy and success rate. In this method, the success rate is 65.3\%, and the precision is 86.59\%, which are higher compared to the other two methods. Additionally, the total number of floating-point operations in this method is 2.017 GFLOPs, and the number of parameters is 7.338 M. The execution speed of this method is recorded at 6.9 fps, which is slightly lower than the other two methods but still remains at an acceptable level.


\begin{table}[t!]
    \centering
    \renewcommand{\arraystretch}{1.1} 
    \setlength{\tabcolsep}{7pt} 
    
    \resizebox{1\linewidth}{!}{
    \begin{tabular}{
        C{3.5cm} | C{2cm} | C{2cm} | C{2cm} C{2cm} C{2cm}
    }
        \toprule
        \multirow{2}{*}{\shortstack{\textbf{Combine Type}}} 
        & \multirow{2}{*}{\textbf{FLOPs}} 
        & \multirow{2}{*}{\textbf{\#Param}} 
        & \multicolumn{3}{C{6cm}}{\textbf{UAV123\char64 10fps}} \\
        & & & \textbf{Succ.} & \textbf{Prec.} & \textbf{Speed} \\
        \midrule
        \textbf{Concat} & 2.017 G & 7.338 M & {\textbf{65.3}} & {\textbf{86.59}} & $6.9$ \textit{fps} \\
        \textbf{Mean} & {\textbf{1.998 G}} & 7.045 M & 64.59 & 85.48 & $6.92$ \textit{fps} \\
        \textbf{Sum} & {\textbf{1.998 G}} & {\textbf{7.042 M}} & 64.5 & 85.9 & \textbf{6.93 \textit{fps}} \\
        \bottomrule
    \end{tabular}
    }
    \caption{The Impact of Different Template Feature Combination  Methods.}
    \label{tab:7}
\end{table}

\noindent\textbf{Evaluation of Different Backbones:} 
The model has been evaluated using different backbones to assess the impact of various architectures on overall performance. All these backbones have been pre-trained on the ImageNet dataset. In this experiment, the following evaluation metrics have been considered: Success Rate, Precision, Floating Point Operations, Number of Parameters, and Execution Speed (Speed).
The results reported in Table \ref{tab:8} on the UAV123@10fps \cite{benchmark2016benchmark} benchmark indicate that the ShuffleNet\_v2 \cite{zhang2018shufflenet} model achieves the best performance among different backbones. This model demonstrates a success rate of 65.3\% and a precision of 86.59\%, outperforming other backbone architectures. Additionally, due to its lowest computational cost (0.199 GFLOPs) and smallest number of parameters (0.766 M), it maintains a high execution speed of 6.9 fps. These characteristics make ShuffleNet\_v2 \cite{benchmark2016benchmark} the most suitable choice for use in the T-SiamTPN model.


\begin{table}[t]
    \centering
    \renewcommand{\arraystretch}{1.1} 
    \setlength{\tabcolsep}{4pt} 
    
    \resizebox{\linewidth}{!}{
    \begin{tabular}{
        C{4.1cm} | C{1.5cm} C{2.5cm} | C{2cm} | C{2cm} | C{1.3cm} C{1.3cm} C{1.3cm}
    }
        \toprule
        \multirow{2}{*}{\textbf{Backbone}} 
        & \multicolumn{2}{C{4.1cm}|}{\textbf{Output Stage}} 
        & \multirow{2}{*}{\textbf{FLOPs}} 
        & \multirow{2}{*}{\textbf{\#Param}} 
        & \multicolumn{3}{C{4.3cm}}{\textbf{UAV123\char64 10fps}} \\
        & \textbf{Stage} & \textbf{Channel} & & & \textbf{Succ.} & \textbf{Prec.} & \textbf{Speed} \\
        \midrule
        \textbf{ResNet18* \cite{he2016deep}} & 2, 3, 4 & 64, 128, 256 & 2.783 G & 2.298 M & 56.35 & 79.35 & $3.3$~\textit{fps} \\
        \textbf{EfficientNet\_b0 \cite{tan2019efficientnet}} & 3, 5, 7 & 40, 112, 320 & 0.599 G & 4.008 M & 51.36 & 69.83 & $6$~\textit{fps} \\
        \textbf{MobileNet\_v2 \cite{sandler2018mobilenetv2}} & 3, 5, 7 & 32, 96, 320 & 2.248 G & 1.813 M & 54.1 & 73.38 & $2.8$~\textit{fps} \\
        \textbf{ShuffleNet\_v2 \cite{zhang2018shufflenet}} & 2, 3, 4 & 116, 232, 464 & \textbf{0.199 G} & \textbf{0.766 M} & \textbf{65.3} & \textbf{86.59} & \textbf{6.9\textit{fps}} \\
        \bottomrule
    \end{tabular}
    }
    \caption{Comparison of different backbone architectures. The ResNet18* model has fewer parameters when the last layer is removed.}
    \label{tab:8}
\end{table}

\section{Real-World Tests}

\begin{figure}[t] 
    \centering
    \includegraphics[width=\columnwidth]{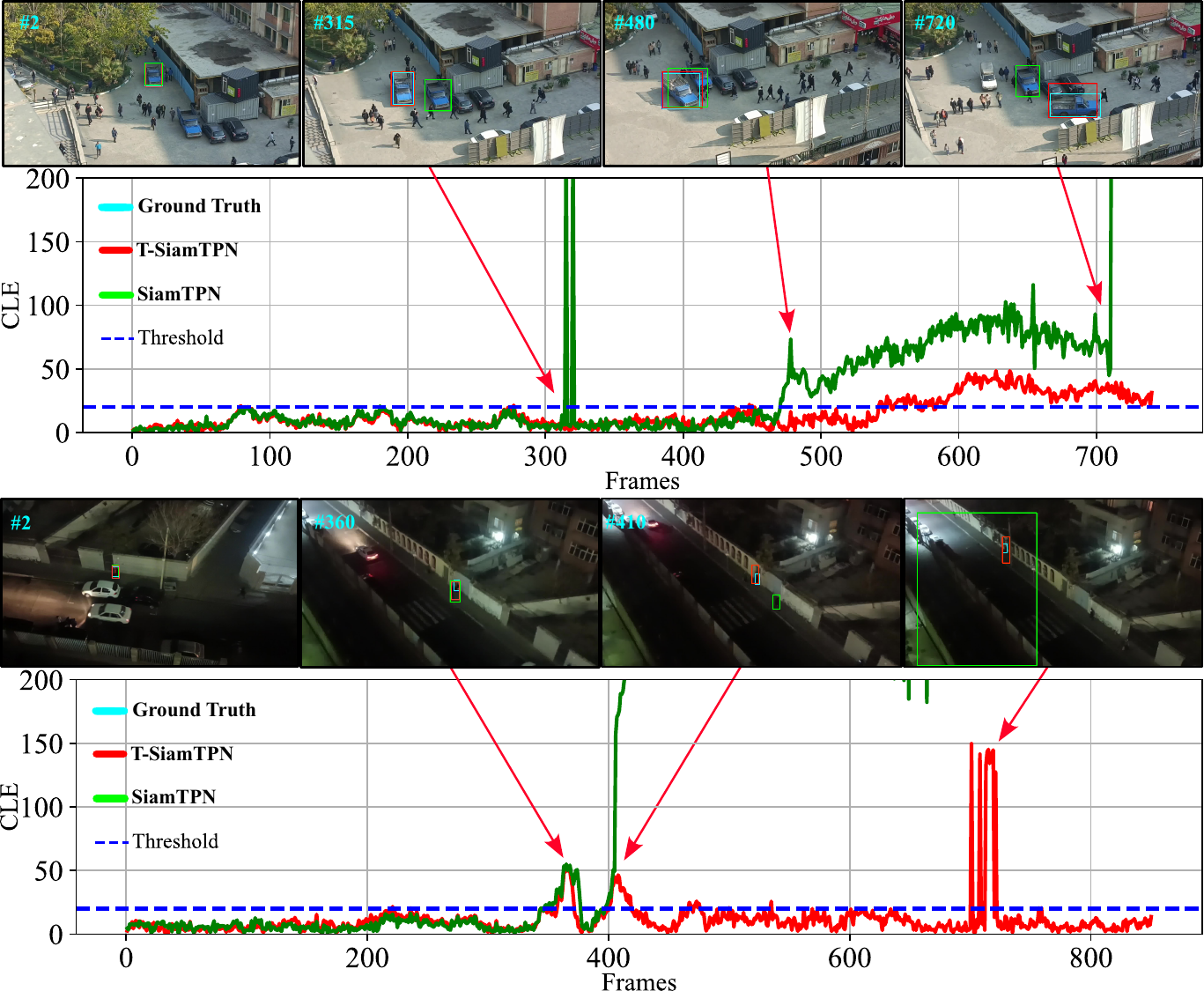} 
    \caption{Tracking performance of T-SiamTPN vs. SiamTPN in real-world day and night scenarios with challenges like occlusion, fast motion, and cluttered backgrounds.}
    \label{fig:9}
\end{figure}

To assess the real-world performance of T-SiamTPN, an experiment is conducted on an NVIDIA Jetson Nano (4GB GPU, 4GB RAM). The test involved two videos recorded at 1080×1920 resolution, one captured during the day and the other at night, featuring challenges such as similar objects, fast camera motion, partial occlusion, low resolution, and background clutter.
As shown in Figure \ref{fig:9}, T-SiamTPN demonstrated more stable and accurate tracking compared to SiamTPN. The accompanying charts illustrate a comparison of both models based on the CLE evaluation metric.
During the experiment, the average system resource usage was recorded as follows: CPU: 79.03\%, GPU: 41.4\%, and RAM: 46.82\%.

\section{Conclusion}

In this work, we presented T-SiamTPN, an enhanced UAV tracking framework that builds upon the original SiamTPN by addressing the specific challenges of aerial tracking. UAV scenarios typically involve small targets, low-resolution imagery due to lightweight onboard cameras, and relatively limited appearance variation from top-down views. These constraints were carefully considered in refining the architecture and designing a smart, adaptive template update mechanism to improve robustness and reduce model drift.
T-SiamTPN integrates temporal information through a lightweight temporal modeling mechanism and replaces conventional cross-correlation with a multi-scale transformer attention module, boosting its ability to discriminate targets in complex scenes. The inclusion of the Modulated Pooling Attention (MPA) block further refines feature representations, particularly for small objects.
One of the key contributions of this work is the adaptive template update strategy, which dynamically validates update conditions such as confidence, displacement, and size variation. This prevents erroneous updates that could confuse the tracker and degrade performance, especially in challenging scenarios with partial or full occlusion. As shown in Table~\ref{tab:3}, the model achieves over 15\% improvement in both success and precision metrics compared to the second-best method under occlusion conditions.
In summary, T-SiamTPN delivers a well-balanced solution for real-time aerial tracking by combining accuracy, stability, and efficiency. Its design makes it a strong candidate for deployment in practical UAV-based tracking systems.

\end{document}